%% file: output.tex
\documentclass[letterpaper, 10 pt, conference]{ieeeconf}  % Comment this line out if you need a4paper
\usepackage{graphicx}
\usepackage{amsmath}
\usepackage{amssymb}
\usepackage{placeins}
\usepackage{subcaption} 
\usepackage{caption}
\usepackage{xcolor}
\usepackage{hyperref}
\usepackage{booktabs}
\usepackage{cite}
\usepackage{makecell} 

\usepackage{multirow}
\usepackage{tabularx}
\usepackage{algorithm}
\usepackage{algorithmicx}
\usepackage{algpseudocode}

\IEEEoverridecommandlockouts                              % This command is only needed if 
\title{\LARGE \bf
Analyzing Fundamental Diagrams of Mixed Traffic Control at Unsignalized Intersections
}

\author{Iftekharul Islam and Weizi Li% <-this % stops a space
\thanks{Iftekharul Islam and Weizi Li are with the Min H. Kao Department of Electrical Engineering and Computer Science at the University of Tennessee, Knoxville, TN, USA {\tt\small mislam73@vols.utk.edu, weizili@utk.edu}}%
\thanks{}
}

\begin{document}

\maketitle
\thispagestyle{empty}
\pagestyle{empty}

%%%%%%%%%%%%%%%%%%%%%%%%%%%%%%%%%%%%%%%%%%%%%%%%%%%%%%%%%%%%%%%%%%%%%%%%%%%%%%%%
\begin{abstract}
This report examines the effect of mixed traffic, specifically the variation in robot vehicle (RV) penetration rates, on the fundamental diagrams at unsignalized intersections. Through a series of simulations across four distinct intersections, the relationship between traffic flow characteristics were analyzed. The RV penetration rates were varied from 0\% to 100\% in increments of 25\%. The study reveals that while the presence of RVs influences traffic dynamics, the impact on flow and speed is not uniform across different levels of RV penetration. The fundamental diagrams indicate that intersections may experience an increase in capacity with varying levels of RVs, but this trend does not consistently hold as RV penetration approaches 100\%. The variability observed across intersections suggests that local factors possibly influence the traffic flow characteristics. These findings highlight the complexity of integrating RVs into the existing traffic system and underscore the need for intersection-specific traffic management strategies to accommodate the transition towards increased RV presence. 

\end{abstract}

%%%%%%%%%%%%%%%%%%%%%%%%%%%%%%%%%%%%%%%%%%%%%%%%%%%%%%%%%%%%%%%%%%%%%%%%%%%%%%%%

\input{sections/intro}
\input{sections/method}
\input{sections/results}
\input{sections/conclusion}

% \clearpage
\bibliographystyle{unsrt}
\bibliography{references}

% \newpage
% \input{sections/appendix}

% \section*{ACKNOWLEDGMENT}
% The preferred spelling of the word ÒacknowledgmentÓ in America is without an ÒeÓ after the ÒgÓ. Avoid the stilted expression, ÒOne of us (R. B. G.) thanks . . .Ó  Instead, try ÒR. B. G. thanksÓ. Put sponsor acknowledgments in the unnumbered footnote on the first page.

\end{document}

%% file: sections/intro.tex
\section{Introduction}
\label{sec:introduction}

Our transportation systems stand on the precipice of a revolutionary transformation, driven by the advent of robot vehicles (RVs). These autonomous agents, with their algorithmic decision-making processes, exhibit driving behaviors that diverge significantly from their human counterparts, and thus hold the potential to reshape traffic dynamics in unprecedented ways~\cite{james2018characterizing}. At the heart of understanding these evolving traffic patterns lie fundamental diagrams (FDs), which describe the intricate relationships between traffic flow, density, and speed. These diagrams have long served as the bedrock for traffic flow theory and management strategies~\cite{greenberg1959analysis,daganzo1997fundamentals}. However, traditional FDs, calibrated to human-driven vehicle behaviors, now face scrutiny regarding their applicability in the emerging paradigm of mixed traffic scenarios that incorporate RVs. 

The recent surge in RV technology development, particularly the refinement of self-driving capabilities, has introduced a new repertoire of vehicle behaviors into the traffic ecosystem. These behaviors are fundamentally distinct from those of human drivers, characterized by their precision, adherence to predefined algorithms, and rapid response times. As the penetration rate of RVs in traffic streams increases, comprehending how these technologies impact traffic flow characteristics becomes not just an academic pursuit but a critical necessity for effective traffic management and forward-thinking infrastructure planning.

Empirical studies and theoretical models have begun to shed light on the multifaceted impacts of RVs on traffic flow. Shi and Li~\cite{shi2021constructing} conducted pioneering field experiments with commercial RVs to construct FDs specifically for RV traffic. They observed that while the shortest RV headway setting demonstrated potential for significantly enhancing road capacity, other settings paradoxically led to capacity decreases compared to human-driven traffic. This dichotomy underscores the complexity of integrating RVs into existing traffic systems and the need for careful calibration of autonomous driving parameters. 

In a similar vein, Makridis et al.~\cite{makridis2024platoon} proposed an innovative method to derive FDs from platoon vehicle trajectories. Their research not only highlighted the capacity-enhancing potential of adaptive cruise control (ACC) systems but also illuminated the stark behavioral contrasts between human drivers and partially automated vehicles. This work provides valuable insights into the transition phase where human-driven and autonomous vehicles coexist, emphasizing the need for adaptive traffic management strategies.

The research landscape extends beyond vehicular traffic to analogous high-density scenarios. Helbing et al.~\cite{helbing2007dynamics} delved into the transitions from laminar to stop-and-go and turbulent flows in crowd movements during disasters. While focused on pedestrian dynamics, this work offers valuable parallels to vehicular traffic, particularly in understanding the conditions that precipitate flow transitions. These insights are crucial for anticipating and mitigating potential safety risks and efficiency bottlenecks as traffic compositions evolve.
Complementing empirical studies, simulation-based research has proven instrumental in unraveling the complexities of mixed traffic flow dynamics. Shang and Stern~\cite{shang2021hybrid} introduced a sophisticated numerical method to estimate the composite FD for mixed human-piloted and automated traffic flow. Their approach provides a macroscopic perspective on traffic behavior across various RV market penetration rates, offering a predictive tool for traffic planners and policymakers.
Further enriching the simulation-based approach, Lu et al.~\cite{lu2018impacts} investigated the impacts of RVs on the urban Macroscopic Fundamental Diagram (MFD) through comprehensive traffic simulations. Their findings underscore the necessity of reassessing established traffic dynamics and control methodologies in light of RV integration. Notably, their research suggests that while RVs can significantly enhance traffic efficiency at high penetration rates, the benefits are less pronounced at lower levels of RV integration. This non-linear relationship between RV penetration and traffic efficiency highlights the complexities involved in transitioning to a mixed traffic environment.

This report aims to extend the current understanding of mixed traffic impacts on FDs, with a  focus on unsignalized intersections---a critical yet often overlooked component of urban traffic networks. By meticulously analyzing FDs derived from simulations across four diverse intersections with varying RV penetration rates, we seek to uncover patterns and draw robust conclusions about the effects of mixed traffic on key traffic flow characteristics.
Our analysis delves into several critical aspects:

\begin{itemize}
    \item The evolution of capacity and critical density as RV penetration increases.
    \item Changes in the shape and characteristics of the FDs across different RV penetration levels.
    \item The emergence of new traffic states or phenomena unique to mixed traffic scenarios.
    \item The impact of intersection geometry and traffic composition on the observed FD changes. 
\end{itemize}

This study is a crucial step towards developing adaptive and effective traffic management strategies. As urban areas grapple with the integration of RVs into existing traffic systems, our findings will provide valuable insights for infrastructure designers, traffic engineers, and policymakers. By anticipating the changes in traffic flow characteristics, stakeholders can proactively develop solutions that maximize the benefits of RV technology while mitigating potential challenges during the transition period.
Moreover, this research lays the groundwork for future studies on more complex traffic scenarios, such as networks of intersections or corridors with varying levels of RV penetration. 

%% file: sections/method.tex
\section{Methodology}
\label{sec:method}

\subsection{Traffic Flow Variables}
The variables which are the building blocks of the fundamental diagrams are speed, density, and flow:

\begin{itemize}
    \item \textbf{Speed (V):} The average speed of vehicles passing through a segment of the road network during a specified time interval.
    
    \item \textbf{Density (k):} The number of vehicles occupying a given length of a lane or roadway at a particular instant in time, typically expressed as vehicles per kilometer.

    \item \textbf{Flow (Q):} The rate at which vehicles pass a reference point on the road, usually expressed as vehicles per hour.
\end{itemize}

The relationship among these variables are given by the following equation:
\begin{equation}
 Q = k.V.
\label{eq1}
\end{equation}

%%%%%%%%%%%%%%%%%%%%%%%%%%%%%%%%%%%%%%%%%%%%%%%%%%%
\subsection{Mixed Traffic Control}\label{subsec1}

The control and coordination of vehicles in mixed traffic environment is a rising  topic~\cite{Poudel2024CARL,Poudel2024EnduRL,Wang2024Intersection,Wang2024Privacy,Villarreal2024Eco,Villarreal2023Pixel,Villarreal2023Chat}. The authors present a decentralized multi-agent reinforcement learning approach to handle mixed traffic at unsignalized intersections. Each RV uses a deep reinforcement learning policy to decide whether to stop or go at the intersection entrance, based on observing the surrounding traffic conditions. The policy is trained with a novel conflict-aware reward function that considers both traffic efficiency and potential conflicts between vehicles inside the intersection. A fail-safe coordination mechanism is implemented to eliminate conflicts between RVs at the intersection entrance.

\subsection{Data Collection and Processing}

The data for this study was collected using the Simulation of Urban MObility (SUMO) platform~\cite{behrisch2011sumo}, a highly versatile tool for simulating vehicular traffic. The network with the four intersections referenced in~\cite{Wang2024Intersection} was used and reconstructed in SUMO for simulation purpose\footnote{\url{https://coloradosprings.gov/}}. Simulations were run for each unsignalized intersection, with RV penetration rates varying from 0\% to 100\% in increments of 25\%. Each simulation was executed three times to ensure the robustness of the data. The raw simulation data was processed to calculate the average speed, density, and flow for each intersection and RV penetration rate using the SUMO API\footnote{\url{https://sumo.dlr.de/docs/Tutorials/FundamentalDiagram.html}}. The data points  obtained from the simulations were approximated with polynomial curve fitting to construct the fundamental diagrams. The diagrams were then analyzed to identify any trends or patterns across the different RV penetration rates.

\begin{figure*}[!ht]
\centering
\includegraphics[width=\linewidth]{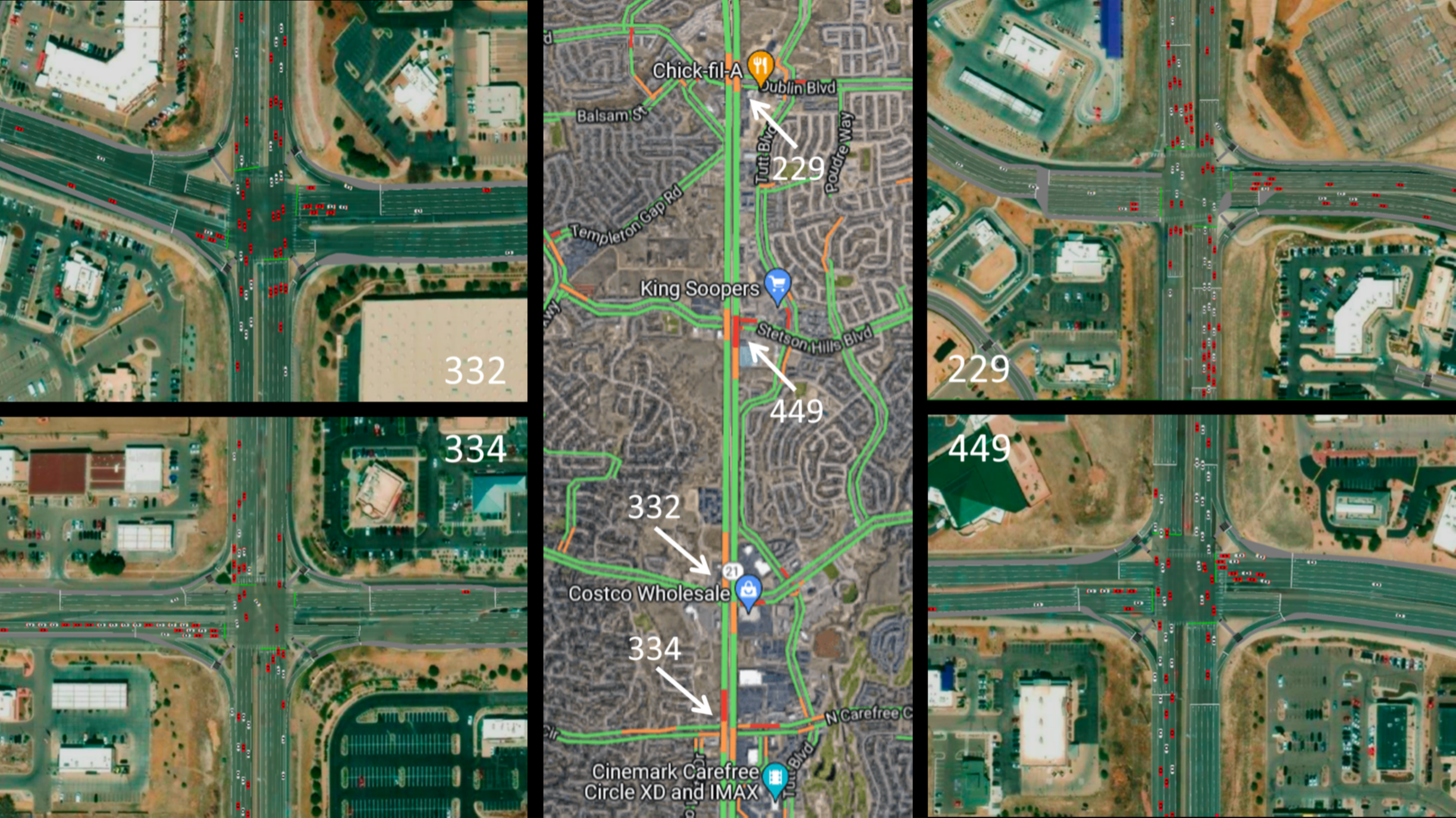}
\caption{\label{fig:intersections}The four intersections in our study.}
\end{figure*}

%% file: sections/results.tex
\section{Simulation Results}
\label{sec:results}

\begin{figure*}[!ht]
\centering
\includegraphics[width=0.9\linewidth]{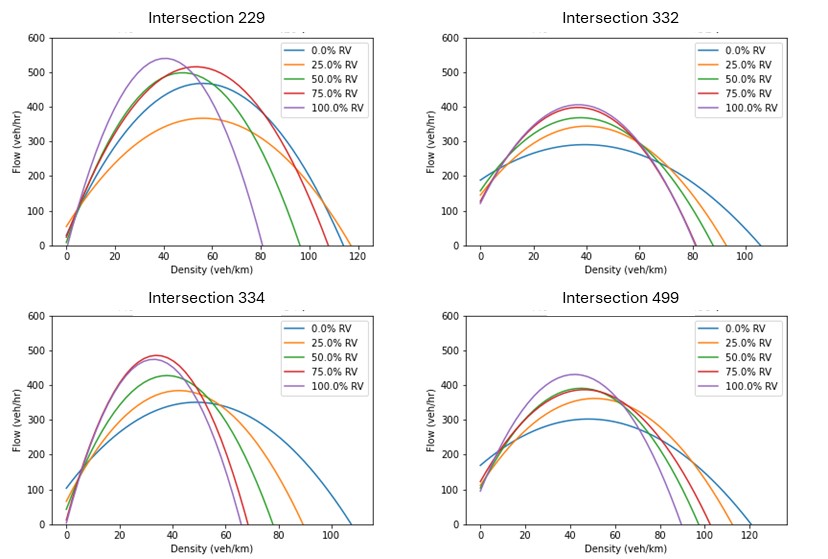}
\caption{\label{fig:flow_density_fig}Fundamental diagrams for the four intersections with different RV penetration rates.}
\end{figure*}

The main results of the simulations are shown in Figure \ref{fig:flow_density_fig} and Table \ref{tab:flow_density_tab}.

\begin{table*}[htb]
  \caption{Polynomial Curve Fitting results ($Q = a.k^{2} + b.k + c$)}  % Replace with your actual title
  \label{tab:flow_density_tab}       % Label for referencing the table
  \centering
  \includegraphics[width=\linewidth]{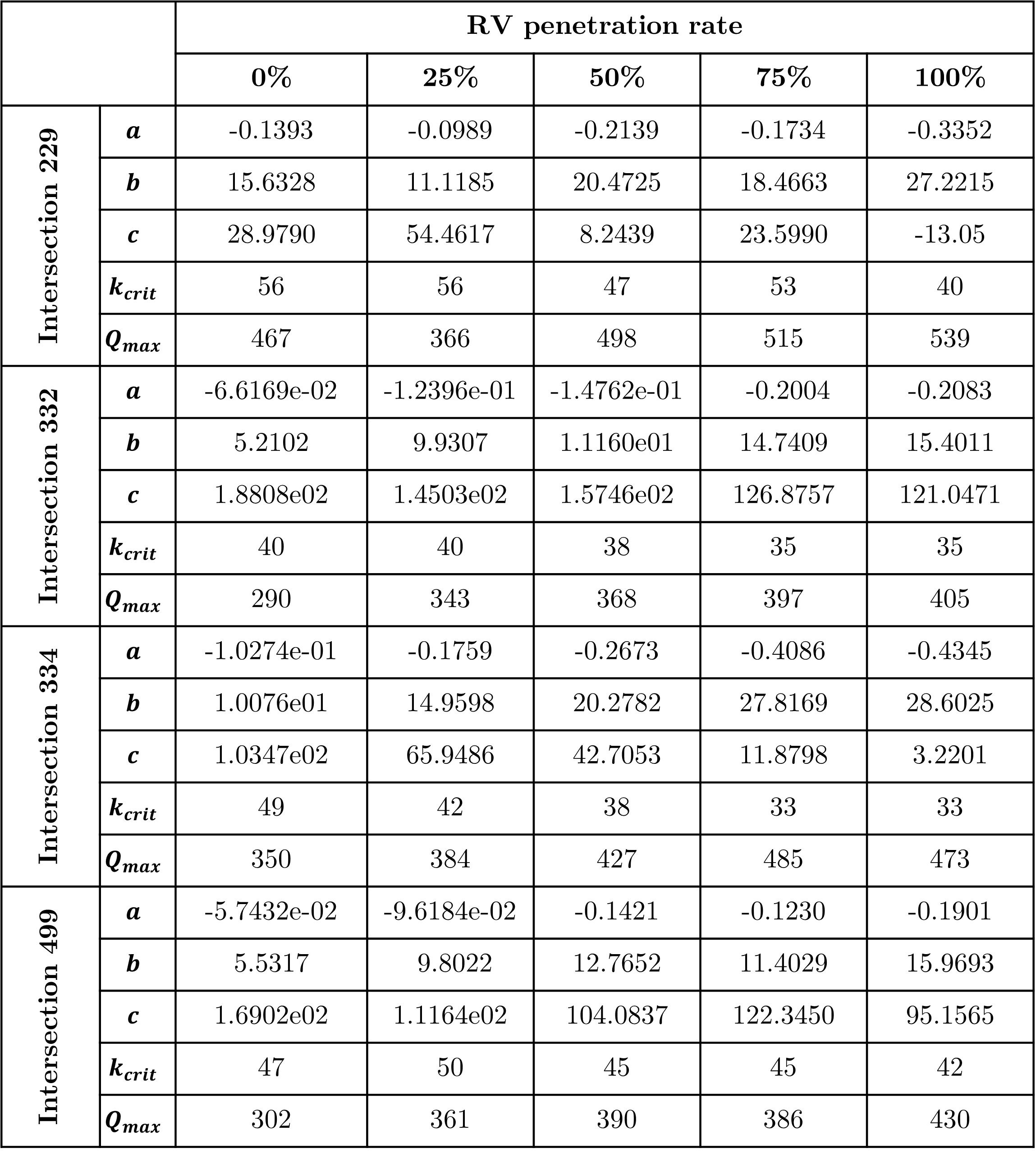}  % Adjust width if needed
\end{table*}

Based on the fundamental diagrams provided for the four intersections (Intersection 229, Intersection 332, Intersection 334, and Intersection 499), we can make the following observations:

\begin{enumerate}
  \item \textbf{Capacity and Congestion:} Each diagram shows a typical flow-density relationship where the flow increases with density up to a certain point ($k_{crit}$), representing the capacity ($Q_{max}$) of the intersection. Beyond this point, the flow decreases as density continues to increase, indicating congestion.
  
  \item \textbf{RV Penetration Impact:} The curves for different RV penetration rates (0\%, 25\%, 50\%, 75\%, and 100\%) show that the flow generally increases with density up to the capacity limit. However, the peak flow (capacity) and the density at which this peak occurs vary slightly with different RV penetration rates.
  
  \item \textbf{Curve Shape:} The shape of the curves is similar across all intersections, suggesting that the fundamental relationship between flow and density is consistent, but the specific values for capacity and the onset of congestion differ.

  \item \textbf{Variability:} There is variability in the flow-density relationship across the intersections, which could be due to differences in intersection design, traffic control measures, or other local factors affecting traffic flow.
  
\end{enumerate}

General conclusions that can be made from these diagrams are:

\begin{itemize}

  \item There is a positive correlation between flow and density up to the capacity limit for all intersections and RV penetration rates.
  \item The presence of RVs affects the flow-density relationship, but the impact varies across different levels of RV penetration and may not be linear.
  \item The capacity and onset of congestion are influenced by the level of RV penetration, but the exact nature of this influence requires further analysis.
  \item The variability observed across intersections suggests that local factors play a significant role in traffic dynamics, and a one-size-fits-all approach may not be suitable for traffic management strategies at unsignalized intersections.
  
\end{itemize}

These observations can inform traffic engineers and planners about the potential impacts of RVs on traffic flow and help guide the development of strategies for managing mixed traffic conditions.

%% file: sections/conclusion.tex
\section{Conclusion and Future Work}
\label{sec:conclude}

The objective of this study was to investigate the effects of increasing RV penetration rates on the fundamental diagrams of traffic flow at unsignalized intersections. The process highlighted the challenges and complexities involved in modeling and understanding mixed traffic systems. In the future, we would like to extend our analysis to larger urban areas, for which city-scale traffic simulation, reconstruction, and estimation techniques can be adopted~\cite{Chao2020Survey,Li2018CityEstIET,Li2017CitySparseITSM,Li2017CityFlowRecon,Wilkie2015Virtual}. 